\documentclass[10pt,conference]{IEEEtran}
\IEEEoverridecommandlockouts
\usepackage{cite}
\usepackage{amsmath,amssymb,amsfonts}
\usepackage{algorithmic}
\usepackage{graphicx}
\usepackage{textcomp}
\usepackage{xcolor}
\usepackage{makecell}
\usepackage{amsmath}
\usepackage{bm}

\def\BibTeX{{\rm B\kern-.05em{\sc i\kern-.025em b}\kern-.08em
    T\kern-.1667em\lower.7ex\hbox{E}\kern-.125emX}}

\usepackage{booktabs}
\usepackage{graphicx}
\usepackage{multirow}
\usepackage{hyperref}

\begin{document}

\title{Multi-dimensional Assessment and Explainable Feedback for Counselor Responses to Client Resistance in Text-based Counseling with LLMs

}



\author{\IEEEauthorblockN{1\textsuperscript{st} Anqi Li}
\IEEEauthorblockA{\textit{Zhejiang University} \\
\textit{Westlake University}\\
Hangzhou, China \\
lianqi@westlake.edu.cn}
\and
\IEEEauthorblockN{2\textsuperscript{nd} Ruihan Wang}
\IEEEauthorblockA{\textit{Westlake University} \\
Hangzhou, China \\
0009-0008-5975-644X}
\and
\IEEEauthorblockN{4\textsuperscript{th} Zhaoming Chen}
\IEEEauthorblockA{\textit{Westlake University}\\
Hangzhou, China \\
0000-0001-7811-6167}
\and
\IEEEauthorblockN{4\textsuperscript{th} Yuqian Chen}
\IEEEauthorblockA{\textit{Westlake University}\\
Hangzhou, China \\
0009-0000-3330-9731}
\and
\IEEEauthorblockN{5\textsuperscript{rd} Yu Lu}
\IEEEauthorblockA{\textit{Westlake University}\\
Hangzhou, China \\
0000-0001-6971-3680}

\and
\IEEEauthorblockN{6\textsuperscript{th*} Yi Zhu}
\IEEEauthorblockA{\textit{Westlake University}\\
Hangzhou, China \\
yuna\_zhu@126.com}
\and
\IEEEauthorblockN{7\textsuperscript{th*} Yuan Xie}
\IEEEauthorblockA{\textit{Westlake University}\\
Hangzhou, China \\
xieyuan@westlake.edu.cn}
\and
\IEEEauthorblockN{8\textsuperscript{th*} Zhenzhong Lan}
\IEEEauthorblockA{\textit{Westlake University}\\
Hangzhou, China \\
lanzhenzhong@westlake.edu.cn}
}

\maketitle

\begin{abstract}

Effectively addressing client resistance is a sophisticated clinical skill in psychological counseling, yet practitioners often lack timely and scalable supervisory feedback to refine their approaches. Although current NLP research has examined overall counseling quality and general therapeutic skills, it fails to provide granular evaluations of high-stakes moments where clients exhibit resistance. In this work, we present a comprehensive pipeline for the multi-dimensional evaluation of human counselors' interventions specifically targeting client resistance in text-based therapy. We introduce a theory-driven framework that decomposes counselor responses into four distinct communication mechanisms. Leveraging this framework, we curate and share an expert-annotated dataset of real-world counseling excerpts, pairing counselor-client interactions with professional ratings and explanatory rationales. Using this data, we perform full-parameter instruction tuning on a Llama-3.1-8B-Instruct backbone to model fine-grained evaluative judgments of response quality and generate explanations underlying. Experimental results show that our approach can effectively distinguish the quality of different communication mechanisms (77-81\% F1), substantially outperforming GPT-4o and Claude-3.5-Sonnet (45-59\% F1). Moreover, the model produces high-quality explanations that closely align with expert references and receive near-ceiling ratings from human experts (2.8-2.9/3.0). A controlled experiment with 43 counselors further confirms that receiving these AI-generated feedback significantly improves counselors' ability to respond effectively to client resistance.

\end{abstract}

\begin{IEEEkeywords}
psychological counseling, text-based, response to resistance, large language models
\end{IEEEkeywords}

\section{Introduction}

\begin{figure}
    \centering
    \includegraphics[width=0.8\linewidth]{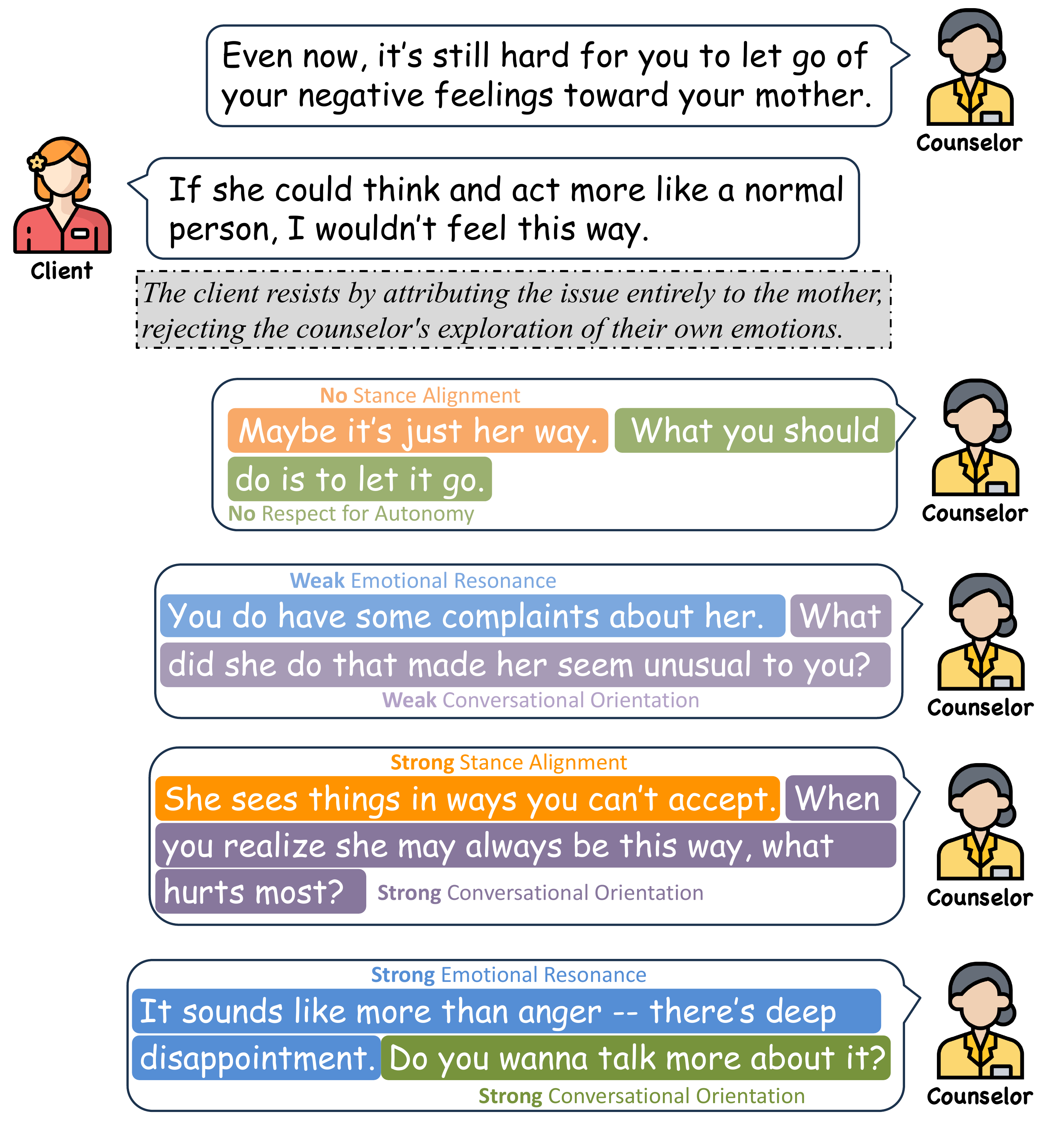}
    \caption{Overview of our framework for evaluating counselor responses to client resistance. The framework comprises four core communication mechanisms: \textit{Respect for Autonomy}, \textit{Stance Alignment}, \textit{Emotional Resonance}, and \textit{Conversational Orientation}. For each mechanism, responses are further categorized into three levels of expression: no expression, weak expression, and strong expression. Our computational approach simultaneously identifies the strength of these dimensions along with their underlying explanations.}
    \label{fig:intro_figure}
\end{figure}

Psychological counseling is a profoundly interactive and dialogue-driven process, the efficacy of which depends largely on the counselor’s capacity to navigate complex interpersonal dynamics. Central to these challenges is the phenomenon of client resistance, where clients oppose or diverge from the established therapeutic trajectory~\cite{beutler2001resistance, newman1994understanding}. Counselors’ responses to such resistance are therefore crucial: when addressed skillfully, it can open a pathway to uncovering clients’ underlying conflicts; when mishandled, it may erode the therapeutic alliance or result in premature termination.

Skillfully responding to client resistance remains a challenge, largely due to the fact that counselors often receive limited and delayed feedback regarding the quality of their interventions. In traditional training and supervision, feedback is typically sparse, retrospective, and dependent on expert availability, making it difficult for counselors, especially novices, to recognize suboptimal responses or explore alternative strategies in a timely manner~\cite{hill1999helping, atkins2014scaling}. As a result, counselors may repeatedly rely on habitual response patterns, even when these patterns inadvertently reinforce resistance. This issue is further exacerbated in text-based counseling settings, where supervision is less structured and many non-verbal interactional cues are absent. Although experienced supervisors can evaluate counselor responses along multiple dimensions, such assessments are neither scalable nor consistently accessible to the broader workforce.

Existing research has explored the feasibility of applying Natural Language Processing (NLP) to provide counselors with feedback~\cite{iyortsuun2023review}. Some studies focus on macro-level analysis, such as classifying overall dialogue quality or understanding global constructs like the therapeutic alliance using LLMs~\cite{li-etal-2024-understanding-therapeutic}. Others concentrate on assessing competencies in specific counseling skills, such as empathy~\cite{sharma2020computational} and reflective listening~\cite{perez2022pair}, by differentiating between high and low quality or providing numeric scores. However, these approaches are generally designed for broad counseling contexts and evaluate discrete, isolated skills. They lack a dedicated framework to evaluate how counselors navigate specific, high-stakes clinical moments, particularly in responding to client resistance.

To address this gap, this study establishes an integrated pipeline spanning theoretical framework, annotated data construction, model development, and empirical validation. In collaboration with psychological experts, and by adapting and synthesizing established theories~\cite{resnicow2012motivational, rosengren2008vaser, di2022RVT}, we propose a four-dimensional framework for evaluating counselor responses to client resistance in text-based counseling. Moving beyond holistic judgments, this framework decomposes counselor replies along four interrelated yet distinct communication mechanisms: respect for autonomy, stance alignment, emotional resonance, and conversational orientation, thereby providing clear and operational criteria for assessment. 

Guided by this framework, we curate a high-quality annotated dataset of resistance-response excerpts drawn from real-world counseling scenarios. Trained psychology students assign multi-dimensional ratings and craft explanatory rationales with high reliability. Leveraging this dataset, we conduct full-parameter instruction tuning on a Llama-3.1-8B-Instruct backbone. Experimental results demonstrate that our model substantially outperforms the zero-shot performance of advanced general-purpose LLMs, including GPT-4o and Claude-3.5-Sonnet, across all four communication mechanisms. Specifically, our model achieves macro-F1 scores of up to 81\%, compared to approximately 45\% for the strongest baseline. Further ablation studies show that incorporating explicit textual explanations during training yields an additional improvement of around 4 F1 points over models trained solely with labels, highlighting the critical role of explanation-based supervision. Beyond classification, our model is also capable of generating high-quality explanatory feedback. Both automated and human evaluations confirm the effectiveness of the generated explanations: our model nearly doubles the BLEU-1 score of top-tier baselines (0.60 vs. 0.32) and receives near-ceiling human ratings (2.8-2.9 out of 3.0) in terms of framework consistency, evidence anchoring, and clarity and specificity. These results indicate that the model not only identifies suboptimal counselor responses but also provides theoretically grounded and actionable guidance. Finally, we examine the practical utility of AI-generated feedback for assisting counselor to address client resistance. In a controlled between–within experiment with 43 counselors, those receiving model-generated feedback show significantly greater improvements in response quality than a control group without LLM-based feedback. These results highlight the potential of our approach as a scalable, theory-informed tool for supporting counselor skill development.

Our framework, dataset, and models will be made available at the following \href{https://anonymous.4open.science/r/SCORE-EVAL-29BD}{URL}.

\section{Related Work}

\noindent\textbf{Computational Analysis in Mental Health Counseling.} Recent studies have increasingly applied NLP techniques to the automated analysis and understanding of large-scale mental health-related conversations~\cite{shatte2019machine}. Some studies adopt a dialogue-level perspective, assessing global constructs such as treatment fidelity~\cite{atkins2014scaling}, the counselor–client therapeutic alliance~\cite{goldberg2020machine}, clients’ personality traits~\cite{khan2020personality, yan2024predictingbigpersonalitytraits}, and mental health states including anxiety and depression symptoms~\cite{iyortsuun2023review}. In parallel, other work conducts fine-grained, process-level analyses to identify counselors’ intervention intents and strategies~\cite{cao2019observing}, as well as clients’ suicide intent~\cite{coppersmith2018natural}, behaviors~\cite{li-etal-2023-understanding}, and emotional states~\cite{tanana2021you, kusal2023systematic}. In addition, research on evaluating counselor performance has largely focused on the quality of specific counseling skills, most notably empathy~\cite{sharma2020computational} and reflective listening~\cite{perez2022pair}.


\noindent\textbf{LLM for Providing Feedback to Counselors.} With the emergence of large language models, a growing body of work has begun to leverage LLMs to provide structured feedback and evaluation to support counselors~\cite{demszky2023using}. Some studies incorporate expert knowledge into LLMs through carefully designed prompts, enabling models to analyze counselor–client dialogues and assess the therapeutic alliance across dimensions. These approaches further identify systematic weaknesses in counselors’ relationship-building abilities and offer targeted guidance for improvement~\cite{li-etal-2024-understanding-therapeutic}.
Beyond alliance-focused analysis,~\cite{chaszczewicz2024multi} introduced a multi-level feedback framework for peer counseling, training a small-scale LLM to infer the counseling goal, identify specific areas for improvement, and propose an alternative response given a specific situation. Similarly,~\cite{hsu2025helping} proposed the CARE system, which diagnoses the counseling strategies required in a given situation and suggests example responses during practice sessions. Through user studies, they demonstrated that CARE effectively supports skill development among novice counselors.

Despite this progress, relatively little work has explicitly addressed client resistance, a critical yet challenging phenomenon in psychological counseling. Recently, ~\cite{li2026recap} developed an automatic model to detect and classify client fine-grained resistance behaviors. In contrast, our work takes a further step by focusing on \textit{the quality of counselors’ responses to client resistance}, aiming to provide more direct and actionable guidance for counselors to adapt and refine their intervention strategies.



\section{Evaluation Framework}

To identify the core dimensions of therapists' moment-to-moment responses that effectively address clients' interpersonal resistance, we propose a novel framework for multidimensional analysis of counselor responses. The framework was developed through close collaboration with professional clinical supervisors who possess extensive experience in both psychotherapy practice and therapist supervision. Drawing on established psychological theories~\cite{di2021rolling, resnicow2012motivational}, we synthesize key dimensions that are central to addressing client resistance in practice and systematically adapt them to the context of text-based counseling.

The proposed framework comprises four core communication mechanisms that collectively offer a comprehensive perspective on how counselors respond to client resistance: \textit{Respect for Autonomy}, \textit{Stance Alignment}, \textit{Emotional Resonance}, and \textit{Conversational Direction}. For each mechanism, counselor responses are evaluated across three levels of expression: (0) \textit{No Expression}, where the mechanism is entirely absent; (1) \textit{Weak Expression}, where it is present but only minimally articulated; and (2) \textit{Strong Expression}, where it is clearly and substantively demonstrated.

We illustrate this framework using the following client resistance scenario: \textit{the counselor invites the client to consider self-care, while the client resists the suggestion by framing rest as incompatible with their perceived family responsibility}.

\noindent\textbf{(1) Respect for Autonomy.} Respect for Autonomy refers to the counselor's recognition and support of the client's self-determination, a foundational principle underlying the entire therapeutic process. A \textbf{no expression} of autonomy respect is marked by didactic responses that pressure the client toward compliance (e.g., \textit{If you burn out, that will hurt your family more—that’s truly irresponsibility}). In contrast, a \textbf{strong expression} affirms the client’s values and encourages exploration of their own thoughts without imposing outcomes (e.g., \textit{Exactly, being the backbone of your family carries a great deal of responsibility}).

\noindent\textbf{(2) Emotional Resonance.} Emotional resonance indicates the counselor’s ability to accurately perceive, understand, and reflect the client’s deep emotional experiences, fostering a sense of containment and psychological safety that is beneficial for the decrease of the resistance behaviors and advancement of therapeutic processes. A \textbf{no expression} neglects or distorts the client's emotions (e.g, \textit{From a family systems perspective, an overfunctioning backbone can foster dependency in others.}). A \textbf{weak expression} reflects superficial empathy, focusing only on external events or rational analysis (e.g., \textit{Yes, having family responsibilities is not easy.}). A \textbf{strong expression} articulates the client's underlying emotions and intentions (e.g, \textit{Carrying that weight alone must be exhausting and lonely at times.}).

\noindent\textbf{(3) Stance Alignment.} Stance alignment refers to the counselor's adjustment of their communication to align with with client's position or perspective. A \textbf{no expression} indicates an oppositional or challenging stance (e.g, \textit{But have you considered that believing you're indispensable may reflect an unbalanced family pattern?}). A \textbf{weak expression} positions the counselor as a detached, third-party observer, acknowledging the situation without fully engaging with the client’s perspective (e.g., \textit{This is a real dilemma.}). A \textbf{strong expression} demonstrates cognitive alignment with the client, supporting their values while collaboratively exploring constructive approaches (e.g, \textit{Your family comes first—that's clear and admirable. Let's figure out how you can keep being their strong support, but in a way that doesn't wear you down over time.})

\noindent\textbf{(4) Conversational Orientation.}
Conversational direction describes how the therapist's response can guide the conversation toward a deeper and a more meaningful exploration of the client's psychological status, alleviating client resistance by fostering a sense of direction within the session. A \textbf{no expression} persuade or debate with the client that lead the counseling process toward a negative trajectory (e.g., \textit{Now we should finish today’s stress assessment form first.}). A \textbf{weak expression} use repetitive responses or aimless questions can contribute to conversational stagnation (e.g., \textit{It’s true.}). A \textbf{strong expression} provides a clear, collaborative path for exploration (e.g., \textit{What does an ideal, well-functioning backbone look like in your mind?}).



\section{Data Collection}

To enable computational evaluation of counselors’ responses to client resistance, we construct a dataset grounded in the proposed framework.

\subsection{Data Source}


 Our data is sourced from two research available counseling dialogue datasets available for research purposes: ClientBehavior~\cite{li-etal-2023-understanding} and ObserverWAI~\cite{li-etal-2024-understanding-therapeutic}. Both datasets consist of text-based counseling conversations between clients and licensed counselors, collected from online mental health platforms in China. To identify client utterances exhibiting resistance, we apply RECAP~\cite{li2026recap}, an automated resistance detection tool with a reported accuracy of 91.41\%. For each detected resistance utterance, we retrieve the counselor’s immediate subsequent response, thereby constructing paired samples of client resistance contexts and corresponding counselor interventions.

\subsection{Annotation Task and Process}

\begin{table}[]
\centering
\caption{Statistics of the collected resistance response evaluation dataset.}
\begin{tabular}{l|ccc|c}
\toprule
\textbf{Dimension} & \textbf{No} & \textbf{Weak} & \textbf{Strong} & \textbf{Total} \\
\midrule
Respect for Autonomy            & 561         & 2761          & 514             & 3836           \\
Stance Alignment                & 445         & 2654          & 737             & 3836           \\
Emotional Resonance             & 1299        & 2262          & 275             & 3836           \\
Conversation Orientation        & 1561        & 1813          & 462             & 3836    \\
\bottomrule
\end{tabular}
\label{tab:data_statistics}
\end{table}

\paragraph{Annotators Recruitment and Training} Prior psychological research has demonstrated that assessing the quality of counselor responses to client resistance is a highly challenging task, one that cannot be reliably performed by crowdworkers~\cite{test_therapist_recognition_correct_response}. To ensure high-quality annotations, we recruited five licensed counselors as annotators, each with over ten years of clinical practice and formal supervisory experience. All annotators received structured training using a detailed annotation manual and framework-specific guidelines before the annotation process.

\paragraph{Conversational Mechanisms Annotation} In our annotation task, annotators were presented with pairs consisting of a counseling dialogue segment ending with a client resistance utterance and the subsequent counselor response. For each such pair, annotators were instructed to assess the counselor response along four predefined conversational mechanisms derived from our proposed framework. Specifically, for each mechanism, they rated whether it was: not expressed, weakly expressed, or strongly expressed within the given context of client resistance.

\paragraph{Explanations Crafting} Following the categorical annotations, annotators were also asked to provide detailed explanations for their labels. Each explanation should first analyze the client’s resistance behavior and then examine the counselor’s response. The analysis must highlight the elements of the response that are critical for the assigned annotation category and provide a natural-language explanation in accordance with the evaluation criteria. For instance, considering the resistance context and the first counselor response in Figure~\ref{fig:intro_figure}, an example explanation is as follows: \textit{Client Resistance Analysis: The client rejects the counselor’s statement regarding their unresolved feelings toward their mother and instead attributes responsibility for their emotional state to the mother’s behavior. Counselor Response Analysis: The counselor’s remark, “Maybe it’s just her way,” functions as a defense of the mother or as minimization of the harm experienced by the client, positioning the counselor in cognitive opposition to the client. Therefore, the response is categorized as No Stance Alignment.}



\paragraph{Quality Control} To ensure annotation reliability, each sample was independently labeled by two annotators. Disagreements were resolved by a third annotator, and the final label was determined via majority vote. The inter-annotator agreement, measured using Cohen's $\kappa$, ranged from 0.74 to 0.77 across the four communication mechanisms, indicating substantial agreement and reflecting the overall quality and consistency of the annotations.

To further assess the quality of the annotated explanations, two taxonomy developers who were not involved in the annotation process to conduct an independent evaluation. A total of 100 samples were randomly selected, and the evaluators reviewed both the annotated communication mechanism levels and the accompanying explanations. Explanations were rated along three dimensions: framework consistency (whether the explanation correctly interprets and applies the core concept of the corresponding framework dimension), evidence anchoring (whether the explanation is explicitly grounded in specific language or behaviors in the counselor’s response), and clarity and specificity (whether the explanation is clearly articulated, concrete, and easy to understand). Each dimension was rated on a 3-point Likert scale. The average scores across the three dimensions were $2.82_{0.38}$, $2.84_{0.37}$, and $2.80_{0.40}$, respectively, indicating that the explanations are of consistently high quality.

The resulting corpus comprises 3,836 samples, each annotated with conversational mechanisms and accompanied by explanations provided by trained annotators.

\paragraph{Privacy and Ethics} Both source datasets are restricted to research use, with all personally identifiable information removed. Our collected data will follow the same protocol, requiring institutions to apply and sign a data-sharing agreement. This work makes no treatment recommendations or diagnostic claims.

\section{Experiment and Results}

Using our collected dataset, we develop a computational approach that leverages large language models to evaluate counselor responses to client resistance.

\subsection{Problem Definition}
For each sample, comprising the dialogue history ending with client resistance and the counselor’s subsequent response, the task is to assess the level of each communication mechanism as \textit{no}, \textit{weak}, or \textit{strong} expression, and to generate free-text explanations justifying these assessments.

\subsection{Experiment Setup}

We employ 5-fold cross-validation to mitigate bias from random data splits and ensure a stable evaluation of model performance. The dataset is partitioned into five folds via stratified sampling, preserving the original distribution across all folds. To address class imbalance, we apply random oversampling on the training set within each fold to balance the sample sizes across different classes. For each fold, four parts are used for training, and the remaining one is held out for validation. We perform full-parameter fine-tuning of the LLaMA-3.1-8B-Instruct model on the training set. Training runs for 3 epochs with an initial learning rate of $1\times10^{-5}$, and is regulated by early stopping based on the validation loss to prevent overfitting. The best model checkpoint from each fold is selected for final evaluation. During inference, we use deterministic decoding with temperature as 0 and top-$p$ as 1.0 to ensure reproducible outputs. The final performance metrics are reported as the mean and standard deviation across all five folds.

\begin{table*}[]
\centering
\caption{Comparison of classification performance between the baseline model and our model on the four communication mechanisms. Results are reported in terms of macro-F1 and accuracy, with the best-performing model in each column shown in bold.}
\scalebox{0.85}{
\begin{tabular}{l|cc|cc|cc|cc}
\toprule
\multirow{2}{*}{\textbf{Model Name}} & \multicolumn{2}{c}{\textbf{Respect for Autonomy}} & \multicolumn{2}{c}{\textbf{Stance Alignment}} & \multicolumn{2}{c}{\textbf{Emotional Resonance}} & \multicolumn{2}{c}{\textbf{Conversational Orientation}} \\
                                     & \textbf{F1}             & \textbf{Acc.}           & \textbf{F1}           & \textbf{Acc.}         & \textbf{F1}             & \textbf{Acc.}          & \textbf{F1}                & \textbf{Acc.}              \\ \midrule
Claude-3.5-Sonnet             & $41.61 \pm  0.99$       & $57.06 \pm  1.56$       & $52.59 \pm  1.41$     & $53.10 \pm  1.09$     & $55.59 \pm  1.44$       & $60.77 \pm  1.21$      & $45.36 \pm  1.62$          & $45.13 \pm  1.97$          \\
GPT-4o                      & $45.37 \pm  1.33$       & $53.31 \pm  1.38$       & $58.61 \pm  1.99$     & $63.87 \pm  1.29$     & $53.16 \pm  1.05$       & $61.52 \pm  1.21$      & $41.72 \pm  0.63$          & $44.26 \pm  0.48$          \\ \midrule
Qwen2.5-7B-Instruct         & $27.92 \pm  1.57$       & $34.91 \pm  1.56$       & $22.97 \pm  1.54$     & $25.34 \pm  1.18$     & $29.34 \pm  1.35$       & $39.49 \pm  1.69$      & $27.33 \pm  1.21$          & $30.63 \pm  1.61$          \\
Qwen2.5-14B-Instruct        & $33.43 \pm  1.71$       & $32.20 \pm  0.64$       & $32.97 \pm  1.29$     & $43.35 \pm  0.93$     & $30.04 \pm  1.38$       & $37.38 \pm  2.06$      & $29.75 \pm  1.04$          & $29.67 \pm  1.17$          \\
Qwen2.5-32B-Instruct        & $34.72 \pm  1.43$       & $44.24 \pm  1.83$       & $37.73 \pm  1.42$     & $47.08 \pm  1.26$     & $33.32 \pm  1.43$       & $38.35 \pm  1.15$      & $30.01 \pm  0.91$          & $29.64 \pm  0.81$          \\
Qwen2.5-72B-Instruct        & $38.90 \pm  1.86$       & $52.16 \pm  1.94$       & $42.10 \pm  2.22$     & $52.16 \pm  1.37$     & $39.24 \pm  0.95$       & $44.92 \pm  0.95$      & $36.62 \pm  0.84$          & $40.74 \pm  0.94$          \\
Llama3.1-8B-Instruct        & $22.13 \pm  0.51$       & $29.20 \pm  1.37$       & $30.55 \pm  1.23$     & $33.97 \pm  1.63$     & $25.57 \pm  1.28$       & $31.25 \pm  0.47$      & $29.37 \pm  1.18$          & $30.74 \pm  1.13$          \\
Llama3.1-70B-Instruct       & $38.06 \pm  1.41$       & $35.79 \pm  0.57$       & $36.62 \pm  1.88$     & $39.11 \pm  0.93$     & $41.38 \pm  1.66$       & $49.06 \pm  1.38$      & $34.46 \pm  0.49$          & $33.91 \pm  0.66$          \\ \midrule
\textbf{Our Model}                   & \bm{$80.92 \pm  1.55$}       & \bm{$87.06 \pm 0.85$}       & \bm{$77.56 \pm  1.78$}     & \bm{$84.06 \pm  1.37$}     & \bm{$77.34 \pm  3.67$}       & \bm{$78.68 \pm  3.32$}      & \bm{$77.87 \pm  0.54$}          & \bm{$77.64 \pm  0.56$}  \\

- Explanations             & $73.24 \pm  1.92$       & $83.38 \pm  0.57$       & $70.17 \pm  1.30$     & $80.74 \pm  0.42$     & $73.23 \pm  2.67$       & $75.52 \pm  3.21$      & $73.21 \pm  1.48$          & $74.15 \pm  1.73$        \\ 
\bottomrule      
\end{tabular}}
\label{tab:main_results}
\end{table*}

\begin{table*}[]
\centering
\caption{Comparison of explanation generation performance between the baseline model and our model. Results are reported in terms of BlEU-1/2, Rouge-1/2/L, and human evaluations, with the best-performing model in each column shown in bold.}
\scalebox{0.86}{
\begin{tabular}{l|ccccc|ccc}
\toprule
\multirow{3}{*}{\textbf{Model Name}} & \multicolumn{5}{c}{\textbf{Automatic Evaluation}}                                                 & \multicolumn{3}{c}{\textbf{Human Evaluation}}                                                                                                                                                                                \\
                                     & \textbf{BLEU-1}   & \textbf{BLEU-2}   & \textbf{Rouge-1}  & \textbf{Rouge-2}  & \textbf{Rouge-L}  & \textbf{\begin{tabular}[c]{@{}c@{}}Framework\\ Consistency\end{tabular}} & \textbf{\begin{tabular}[c]{@{}c@{}}Evidence\\ Anchoring\end{tabular}} & \textbf{\begin{tabular}[c]{@{}c@{}}Clarity \&\\ Specificity\end{tabular}} \\ \midrule
                                     
Claude-3.5-Sonnet                    & $32.45 \pm  0.39$ & $20.45 \pm  0.18$ & $47.58 \pm  0.08$ & $19.02 \pm  0.11$ & $29.14 \pm  0.05$ & $1.94 \pm  0.73$                                                         & $1.94 \pm  0.73$                                                      & $2.18 \pm  0.48$                                                          \\
GPT-4o                               & $24.87 \pm  0.51$ & $14.23 \pm  0.26$ & $42.46 \pm  0.16$ & $14.70 \pm  0.14$ & $25.13 \pm  0.09$ & $2.04 \pm  0.82$                                                         & $2.00 \pm  0.80$                                                      & $2.40 \pm  0.49$                                                          \\ \midrule
Qwen2.5-7B-Instruct                  & $17.94 \pm  0.61$ & $9.99 \pm  0.31$  & $39.20 \pm  0.09$ & $12.46 \pm  0.07$ & $21.69 \pm  0.15$ & $1.90 \pm  0.73$                                                         & $1.90 \pm  0.73$                                                      & $2.26 \pm  0.48$                                                          \\
Qwen2.5-14B-Instruct                 & $6.22 \pm  0.67$  & $3.45 \pm  0.37$  & $30.82 \pm  0.38$ & $8.58 \pm  0.18$  & $15.85 \pm  0.24$ & $2.08 \pm  0.66$                                                         & $2.00 \pm  0.60$                                                      & $2.32 \pm  0.47$                                                          \\
Qwen2.5-32B-Instruct                 & $15.52 \pm  0.20$ & $9.00 \pm  0.11$  & $40.28 \pm  0.16$ & $13.54 \pm  0.06$ & $22.65 \pm  0.07$ & $2.22 \pm  0.58$                                                         & $2.18 \pm  0.55$                                                      & $2.34 \pm  0.47$                                                          \\
Qwen2.5-72B-Instruct                 & $18.49 \pm  0.54$ & $10.66 \pm  0.32$ & $40.54 \pm  0.20$ & $13.94 \pm  0.15$ & $23.50 \pm  0.20$ & $2.10 \pm  0.64$                                                         & $2.08 \pm  0.63$                                                      & $2.28 \pm  0.45$                                                          \\
Llama-3.1-8B-Instruct                & $7.30 \pm  0.23$  & $4.09 \pm  0.13$  & $32.87 \pm  0.23$ & $10.28 \pm  0.13$ & $19.11 \pm  0.15$ & $1.46 \pm  0.61$                                                         & $1.52 \pm  0.64$                                                      & $1.92 \pm  0.44$                                                          \\
Llama-3.1-70B-Instruct               & $4.58 \pm  0.12$  & $2.60 \pm  0.06$  & $30.91 \pm  0.17$ & $9.42 \pm  0.08$  & $18.61 \pm  0.06$ & $1.54 \pm  0.64$                                                         & $1.52 \pm  0.64$                                                      & $2.04 \pm  0.40$                                                          \\ \midrule
\textbf{Our Model}                            & \bm{$60.34 \pm  0.33$} &\bm{$43.39 \pm  0.30$} & \bm{$63.08 \pm  0.14$} & \bm{$33.71 \pm  0.15$} & \bm{$41.09 \pm  0.20$} & \bm{$2.78 \pm  0.58$} & \bm{$2.73 \pm  0.66$} & \bm{$2.88 \pm  0.33$}    \\
\bottomrule
\end{tabular}}
\label{tab:exp_eval}
\end{table*}

\subsection{Baselines}

We additionally evaluate a range of advanced LLMs under zero-shot prompting. The evaluation includes leading closed-source models such as GPT-4o~\cite{2023gpt4} and Claude-3.5-Sonnet~\cite{Anthropic2024}, as well as open-source models from the Qwen2.5-Instruct family (7B, 14B, 32B, and 72B)~\cite{qwen2.5} and the Llama-3.1-Instruct family (8B and 70B)~\cite{llama3modelcard}. All baseline models are evaluated under a unified protocol.


\subsection{Ablation Study}
We evaluated the contribution of explanation supervision to model performance through an ablation study. Specifically, we fully fine-tuned the Llama-3.1-8B-Instruct backbone using only the dimension-level ratings, without incorporating the corresponding explanations during training. To maintain task consistency between training and evaluation, the model was configured to predict only dimension scores, without generating explanations, at test time. The learning rate was set to $5\times10^{-6}$, with all other experimental settings kept identical to the main experiment.



\subsection{Results}

We analyze the effectiveness of our computational approach in evaluating counselors’ responses to client resistance, together with the quality of the generated explanations. Table~\ref{tab:main_results} reports the classification performance (macro-F1 and accuracy), while Table~\ref{tab:exp_eval} presents both automatic evaluation metrics (BLEU-1, BLEU-2, ROUGE-1, ROUGE-2, and ROUGE-L) and human evaluation results (framework consistency, evidence anchoring, and clarity and specificity) for the explanations generated by baseline models and our proposed approach across the four communication mechanisms.

\paragraph{\textbf{Communication Mechanisms Identification Task}} \textbf{Overall, general-purpose LLMs struggle to accurately differentiate among the three levels of communication mechanisms.} Advanced closed-source models, including GPT-4o and Claude-3.5-Sonnet, consistently achieve state-of-the-art performance across all four dimensions, significantly outperforming open-source counterparts with smaller parameter sizes. Among the evaluated dimensions, \textit{Conversational Orientation} emerges as the most challenging for existing models: even the strongest baseline models fail to exceed approximately 45\% in both macro-F1 and accuracy. 

In contrast, our approach achieves substantial improvements across all four communication dimensions, with macro-F1 scores ranging from 77\% to 81\%. Relative to the best-performing general-purpose LLMs, our model yields gains of more than 20 F1 points overall, and up to 35.5 F1 points in the \textit{Respect for Autonomy} dimension. Further per-class analysis indicates that these improvements are primarily driven by a significant increase in recall for the \textit{No expression} category. This capability is particularly critical for real-world applications, as accurately identifying counselor responses that exhibit severely deficient performance is essential for providing effective feedback and preventing the reinforcement of suboptimal counseling interventions.

\textbf{Fine-tuning on task-specific data yields substantial gains, and explanation-augmented training further enhances performance.} Empirical results demonstrate that fine-tuning the backbone model on our proposed task-specific dataset leads to a significant improvement in classification performance across all communication dimensions. In contrast, the base Llama-3.1-8B-Instruct model attains relatively low F1 scores, ranging from 22.13 to 30.55. After fine-tuning, the model achieves a remarkable 2-4× increase in performance. For example, in the \textit{Respect for Autonomy} dimension, the F1 score rises sharply from 22.13 to 80.92, indicating a substantial enhancement.

Moreover, our findings underscore the additional value of explanation-based supervision. Compared with models trained solely on numerical scores, those incorporating textual explanations achieve higher performance across all four communication mechanisms, with improvements of at least 4 F1 points. This result suggests that integrating explicit rationales provides complementary supervisory signals beyond labels alone, enabling the model to better learn the underlying communicative principles.

\paragraph{\textbf{Explanation Generation Task}} The performance of our model significantly surpasses all baseline models across both automatic and human evaluation metrics. In terms of automatic lexical overlap, our model achieves a BLEU-1 score of 0.60, nearly doubling the performance of the strongest baseline, Claude-3.5-Sonnet (0.32). These results indicate that the explanations generated by our model maintain an exceptionally high level of lexical alignment and semantic consistency with the gold-standard human references.

Beyond surface-level metrics, human expert evaluations further confirm the model’s superiority across all three dimensions. Our model consistently receives near-perfect ratings, ranging from 2.8 to 2.9, which markedly outperforms the scores of GPT-4o and Claude-3.5-Sonnet. This superior performance demonstrates the model’s robust capability to generate high-quality, actionable feedback that is both theoretically grounded and contextually precise, highlighting its significant potential as a supportive tool for counselors.


\section{Proof-of-concept Study with Counselors}
We further examine whether real-time AI-based scoring and feedback can support counselors in adjusting their responses to client resistance in a more appropriate and effective manner.



We adopted a mixed between–within experimental design. A total of 43 novice counselors were recruited and randomly assigned to either a control group ($N=21$) or an experimental group ($N=22$). Participants in both groups first completed a pre-phase, during which they were asked to respond to 10 client utterances exhibiting resistance. They then proceeded to a post-phase, in which they revisited the same counseling contexts and revised their original responses from the pre-phase. During the post phase, participants in the control group received no feedback, whereas those in the experimental group were provided with AI-generated evaluation scores and explanatory feedback on their pre-phase responses.

\begin{figure}
    \centering
    \includegraphics[width=0.8\linewidth]{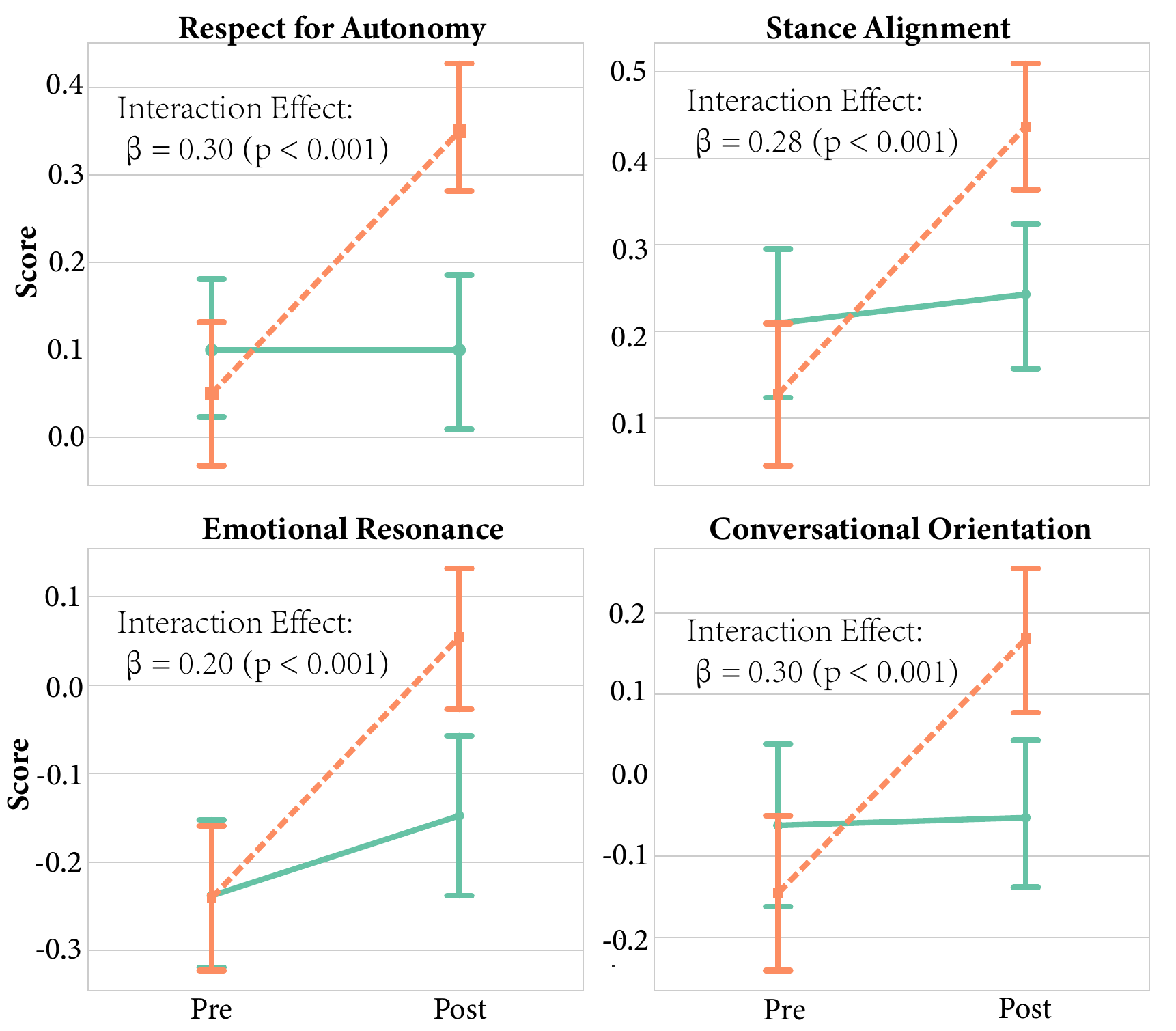}
    \caption{Interaction effects between experimental groups and phases across four dimensions. Solid green lines represent the control group, while dashed orange lines represent the experimental group. Points denote the mean values, and error bars indicate 95\% confidence intervals. The results reveal a significant performance surge in the Experimental Group across all dimensions following the intervention, whereas the Control Group maintains a stable baseline.}
    \label{fig:interaction_effect}
\end{figure}

All counselors' responses were subsequently evaluated using our proposed model to obtain multi-dimensional quality scores. We employed Linear Mixed-Effects Models (LMMs) for each dimension to account for the nested structure of repeated measures within participants. The models included Condition (Experimental vs. Control) and Phase (Pre-test vs. Post-test) as fixed main effects, along with their interaction term. As illustrated in Figure~\ref{fig:interaction_effect}, the analysis reveals a statistically significant Condition $\times$ Phase interaction across all communication mechanisms. While both groups performed comparably at baseline, the interaction term confirms that counselors in the experimental group achieved substantially greater gains in the post-test phase than those in the control group. This robust interaction underscores the efficacy of AI-driven feedback in refining counseling interventions.

Post-experimental semi-structured interviews with the experimental group further validated these quantitative gains. Participants evaluated the system across three core dimensions using a 5-point Likert scale: (1) Increasing awareness of areas needing improvement ($M=4.38, SD=0.65$); (2) Providing clear directions for response refinement ($M=4.14, SD=0.56$); and (3) Enhancing confidence in managing resistant clients ($M=3.86, SD=0.89$). Thematic analysis of participants' free-text reflections revealed several key benefits. Nine participants noted that the feedback allowed for the timely identification of specific weaknesses and actionable improvement paths. Seven participants emphasized that the multi-dimensional nature of the feedback fostered more comprehensive and well-rounded responses. Furthermore, many expressed a strong interest in integrating the system into their ongoing reflective practice, particularly during moments of clinical uncertainty. Regarding future enhancements, participants suggested the inclusion of concrete response exemplars, which represents a promising avenue for subsequent system iterations.



\section{Conclusion}
We propose a novel framework, dataset, and computational approach for distinguishing the multi-dimensional quality of counselor responses to client resistance in text-based counseling. Our method enables fine-grained evaluation of counselors’ responses while providing interpretable explanations for its judgments. Furthermore, through a proof-of-concept study, we demonstrate that delivering AI-generated evaluative feedback to counselors can effectively improve their ability to respond to client resistance.

\section*{Ethics Statement}

\textbf{Data Privacy and Release.} All experts involved in the annotation process were bound by strict confidentiality agreements, and the data were stored on secure, encrypted servers with access restricted to authorized personnel only. The annotated dataset will be released exclusively for research purposes, with access granted only upon submission and approval of a data use application.

\textbf{Ethical Oversight.} The study protocol was reviewed and approved by the Institutional Review Board (IRB) of our university (number withheld for anonymous review). Senior counselors were recruited as annotators with informed consent and were fairly compensated at a rate of approximately 60-80 RMB per hour. Participants involved in the proof-of-concept study also provided informed consent and received compensation of 40 RMB.

\textbf{Intended Use of Automatic Feedback.} While our model demonstrates strong performance in evaluating responses, we emphasize that it is not designed as a replacement for professional clinical judgment. In our controlled experiment, participants were explicitly informed that the feedback was AI-generated and were instructed to use it as a supplemental resource.






\bibliographystyle{IEEEtran}
\bibliography{custom}
%




\end{document}